\documentclass[10pt,twocolumn,letterpaper]{article}

 \usepackage{cvpr}              

\usepackage{ragged2e} 
\usepackage{booktabs,makecell, multirow, tabularx}
\usepackage{pifont}											
\usepackage{colortbl}  
\definecolor{pinkkk}{RGB}{237,218,244}
\definecolor{pinkkk1}{RGB}{248,235,249}
\definecolor{grayyy}{RGB}{242,242,242}
\definecolor{greenn}{RGB}{124,191,51}
\usepackage{array}   
\usepackage{subcaption} 
\usepackage{float}		
\usepackage{algorithm}
\usepackage{algpseudocode}
\usepackage{bm}
\usepackage{extpfeil}
\usepackage{tikz}

\definecolor{cvprblue}{rgb}{0.21,0.49,0.74}
\usepackage[pagebackref,breaklinks,colorlinks,citecolor=cvprblue]{hyperref}

\def\paperID{3068} 
\def\confName{CVPR}
\def\confYear{2025}

\title{SignEye: Traffic Sign Interpretation from Vehicle First-Person View}

\author{Chuang Yang\\
HKPU\\
{\tt\small c1yang@polyu.edu.hk}
\and
Xu Han\\
NWPU\\
\and
Tao Han\\
HKUST\\
\and
Yuejiao Su\\
HKPU\\
\and
Junyu Gao\\
TeleAI\\
\and
Hongyuan Zhang\\
HKU\\
\and
Yi Wang\\
HKPU\\
\and
Lap-pui Chau\\
HKPU\\
}

\begin{document}
\maketitle
\begin{abstract}
Traffic signs play a key role in assisting autonomous driving systems (ADS) by enabling the assessment of vehicle behavior in compliance with traffic regulations and providing navigation instructions. However, current works are limited to basic sign understanding without considering the egocentric vehicle's spatial position, which fails to support further regulation assessment and direction navigation. Following the above issues, we introduce a new task: traffic sign interpretation from the vehicle's first-person view, referred to as \textbf{TSI-FPV}. Meanwhile, we develop a traffic guidance assistant (\textbf{TGA}) scenario application to re-explore the role of traffic signs in ADS as a complement to popular autonomous technologies (such as obstacle perception). Notably, TGA is not a replacement for electronic map navigation; rather, TGA can be an automatic tool for updating it and complementing it in situations such as offline conditions or temporary sign adjustments. Lastly, a spatial and semantic logic-aware stepwise reasoning pipeline (\textbf{SignEye}) is constructed to achieve the TSI-FPV and TGA, and an application-specific dataset (\textbf{Traffic-CN}) is built. Experiments show that TSI-FPV and TGA are achievable via our SignEye trained on Traffic-CN. The results also demonstrate that the TGA can provide complementary information to ADS beyond existing popular autonomous technologies.
\end{abstract}    
\begin{figure}[t]
	\centering
	\includegraphics[width=0.9\linewidth]{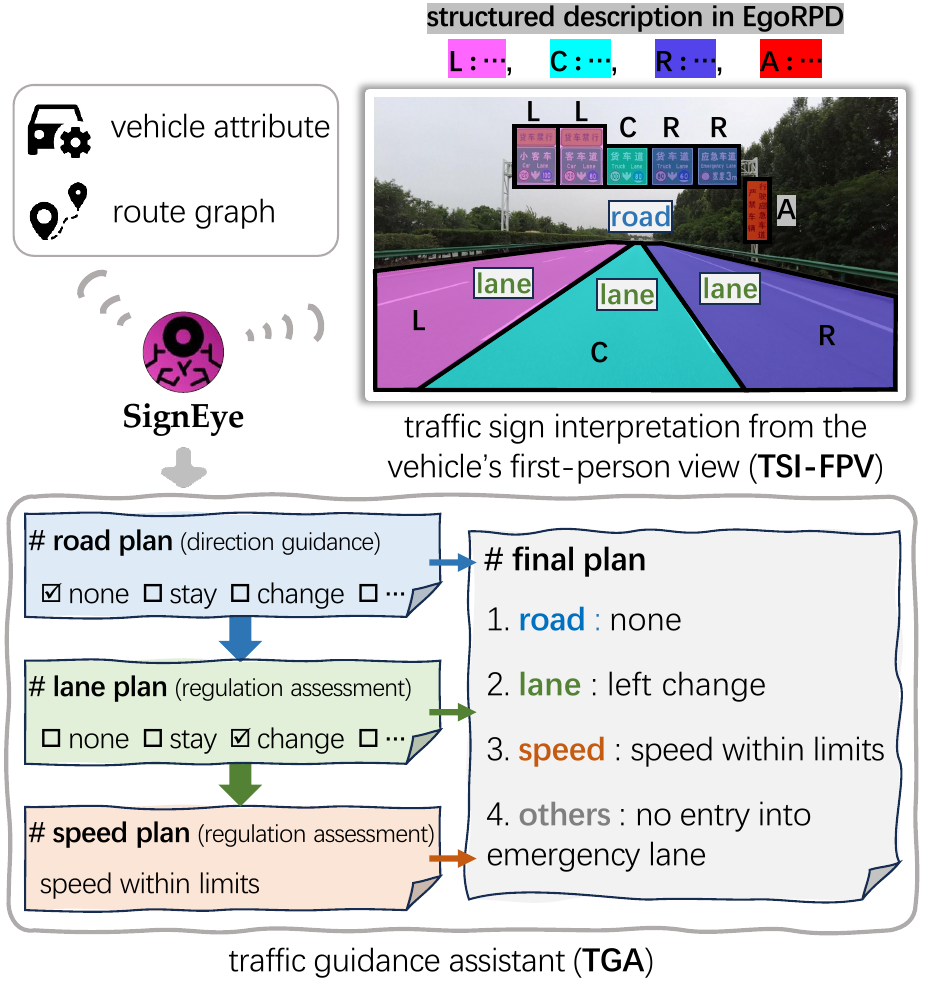}
	\vspace{-.1in}
	\caption{Illustration of the SignEye, where the TSI-FPV part interprets sign units as structured descriptions and assigns them to different roads and lanes in egocentric relative position definition (EgoRPD), where "C" for current, "L" for left, "R" for right, and "A" for all lanes or roads. The TGA part combines the descriptions with vehicle attributes and a route graph to achieve traffic regulation assessment and direction navigation.}
	\label{fig:V1}
	\vspace{-.15in}
\end{figure}
\section{Introduction}
\label{sec:intro}
Vision-based autonomous driving requires perceiving the road environment around vehicles to support decision-making regarding driving plans. Traffic signs (including guide panels, symbols, and text), key components of the road scene, contain traffic regulation and navigation information, which helps formulate a driving plan that adheres to road driving criteria. Traditional traffic sign-related methods primarily focus on the basic traffic symbol (such as the speed limitation and no parking symbols) detection and recognition~\cite{zhu2016traffic,de1997road}. Although recent approaches~\cite{guo2021learning,guo2023visual,yang2024traffic} attempt to combine traffic symbols and texts for comprehensive understanding, the lack of consideration for the egocentric vehicle's spatial position makes them hard to establish a connection between sign information and the vehicle. It leads to current methods still struggling to support autonomous driving systems (ADS) for traffic regulation assessment and direction navigation. 
\begin{figure*}[t]
	\centering
	\includegraphics[width=0.82\linewidth]{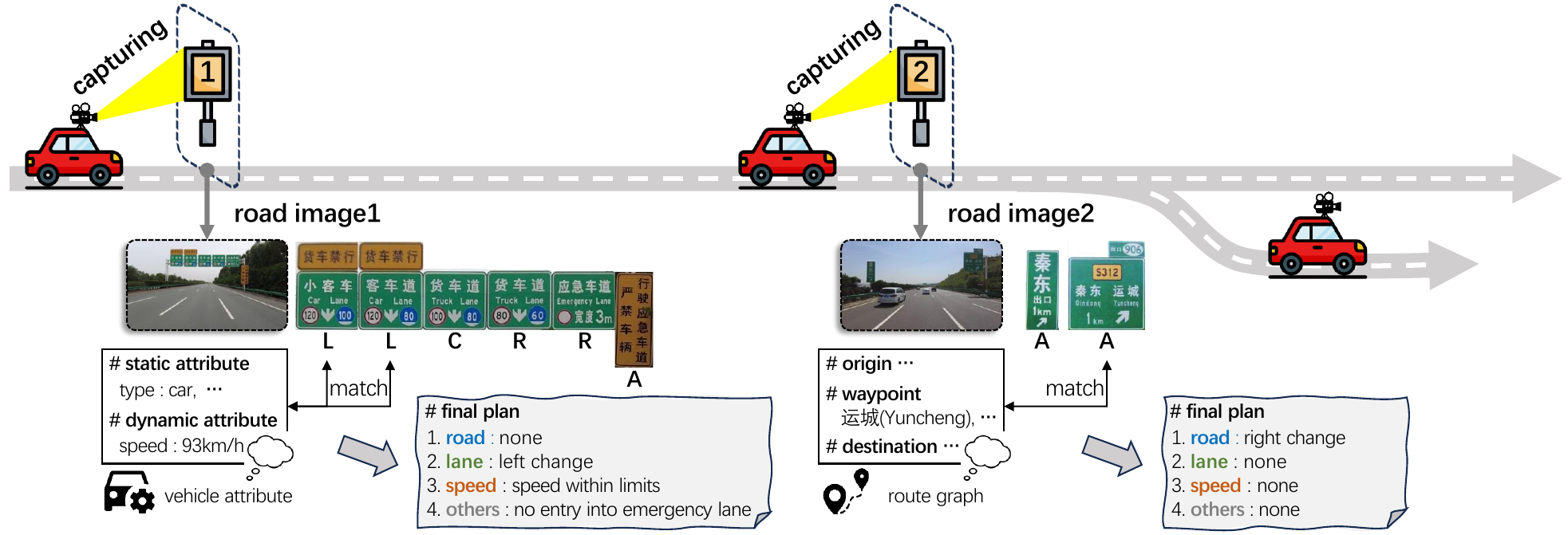}
	\vspace{-.1in}
	\caption{TGA estimates the structured sign descriptions in EgoRPD the vehicle attribute and route graph for achieving traffic regulation assessment and direction navigation respectively.}
	\label{fig:V2}
	\vspace{-.15in}
\end{figure*}

Based on the above observations, we propose to interpret traffic signs from the vehicle's first-person view (TSI-FPV) to assist ADS in decision-making. Specifically, as shown in Figure~\ref{fig:V1}, TSI-FPV \textbf{firstly} generates a description for each traffic \textbf{sign unit} (a sign set, which includes multiple symbols and texts that need to be combined together to interpret complete traffic information) according to the~\href{https://openstd.samr.gov.cn/bzgk/gb/newGbInfo?hcno=15B1FC09EE1AE92F1A9EC97BA3C9E451}{Road Traffic Signs and Markings Criteria}. The interpretation process from a sign unit to a description organizes semantic logic among signs within a unit accurately and naturally. It ensures reliable sign-understanding results. Meanwhile, the description-styled output enjoys intuitional and structured sentence patterns, which makes it easier for ADS or human drivers to take in to further assist in driving plan decision-making. TSI-FPV \textbf{then} distinguishes road and lane markings from the vehicle's first-person view. Lanes within one road are labeled as left (`L'), current (`C'), and right (`R'), where the label `C' indicates the vehicle's located position from an egocentric view, label `L' and `R' are defined according to the determined label `C'. Similarly, roads are labeled in the same way. Compared with the normal absolute strategy (determining the positional order of all lanes or roads from left to right), the introduced egocentric relative position-definition (EgoRPD) strategy only requires considering lanes around the vehicle, which avoids the recognition interference brought by perspective distortion of the distant lane and can provide adequate adjacent lane information for single-step lane change (a recommended practice under~\href{https://flk.npc.gov.cn/detail2.html?ZmY4MDgxODE3YWIyMzFlYjAxN2FiZDYxN2VmNzA1MTk}{Road Traffic Safety Law}) decision-making process at the same time. \textbf{Lastly}, TSI-FPV distributes the descriptions of all sign units in the image to different roads and lanes based on the analysis of the spatial relationship between units, roads, and lanes and the directional information involved in the descriptions. Notably, on the current road, a description is labeled as `A' if the corresponding sign unit regulates all lanes instead of a specific one lane.

Meanwhile, to re-explore the role of traffic signs in ADS, a traffic regulation assistant (TGA) is developed in this paper (as visualized at the bottom of Figure~\ref{fig:V2}). TGA introduces vehicle attribute and route graph into the driving plan decision-making process, where the vehicle attribute consists of dynamic (vehicle speed) and static attributes (the type, size, and weight of the vehicle), and the route graph contains all latent waypoints along the travel route from origin to destination. In decision-making, TGA estimates the information between TSI-FPV output and the vehicle attribute and route graph for achieving traffic regulation assessment (e.g., speed limitations, lane restrictions, distance keeping, etc.) and direction navigation respectively.

Furthermore, to achieve TGA, a stepwise reasoning pipeline (SignEye) is constructed, which decomposes a complex spatial and semantic logic reasoning task into a series of intuitive problems. With decomposition problems, SignEye analyzes semantic logic among signs more accurately to generate structured descriptions for the TSI-FPV task without a hallucination problem. It also can understand the spatial logic between the descriptions and lane and road more easily for further reasoning the driving plan in the TGA. Meanwhile, compared with the previous three steps (involving detection, classification, and natural language processing)-based traffic sign interpretation methods, SignEye avoids heavy dependence on a complex training process and plenty of difficult-to-obtain intensive symbol and text ground-truth data. It combines these three steps into a single process, which considers sign interpretation as image description and can be directly trained using the description of sparse sign units. Besides, considering the absence of a large available dataset to train SignEye, we build a semi-automatic data engine to generate TSI-FPV and TGA corresponding datasets, namely Traffic-CN, to fulfill the research and evaluation in this field. The contributions of this work can be summarized as follows:

\begin{itemize}
	\item[1.] The traffic sign interpretation task from the vehicle's first-person view (TSI-FPV) is introduced to provide sign descriptions with egocentric positional information for ADS. Additionally, a TSI-FPV-based traffic guidance assistant (TGA) scenario is developed to re-evaluate the role of traffic signs in supporting ADS decision-making for driving plans.

	\item[2.] A stepwise reasoning pipeline, called SignEye, is constructed to improve semantic logic analysis among signs and to determine spatial relationships between sign units, lanes, and roads, effectively facilitating TSI-FPV and TGA. Notably, it simplifies the previously multi-step sign interpretation process into a single-step image description, allowing it to be trained directly on sparse descriptions without requiring intensive symbol and text boxes.
	
	\item[3.] The TSI-FPV and TGA corresponding dataset (Traffic-CN) is built. It fulfills the research evaluation and scheme verification in the field of applying traffic signs in ADS. Meanwhile, Traffic-CN will promote the development of the traffic sign community. 
\end{itemize}
\section{Related Works}
\label{sec:rela}
SignEye is constructed based on sign-related expert models and vision-language models (VLMs). In this section, we review previous methods to provide a clearer understanding of the framework and its advantages.

\subsection{Traffic Sign-Related Method}
Previous works on traffic signs primarily pay attention to basic symbol detection and recognition. Initially, deep object detection frameworks~\cite{redmon2016you,ren2015faster,he2017mask} made the task of locating symbols from scene images and recognizing them~\cite{zhu2016traffic,tabernik2019deep,yang2015towards} achievable. However, traffic signs always convey complex instructions via the combination of symbols and texts with different layouts and color styles, which would lead to information omission and even misunderstanding (e.g., the combination of speed limitation symbol and lane information) if relying merely on symbols. 

Following the above considerations, recent works~\cite{guo2021learning} attempted to achieve a comprehensive sign understanding. Specifically, Guo~\etal~\cite{guo2021learning,guo2023visual,yang2024traffic} detected and recognized symbols and texts by optical character recognition models~\cite{yang2022cm,ye2023deepsolo,yang2023text}. Then symbols and texts were combined according to the analysis of the corresponding layout. Further, the authors proposed a traffic knowledge graph in~\cite{guo2023visual,guo2024signparser}, where roads and lanes were considered in the understanding process. To describe signs more intuitively, Yang~\etal~\cite{yang2024traffic} formulated the traffic knowledge graph as a natural language description. Although these methods achieve a comprehensive sign understanding, the lack of consideration for the egocentric vehicle's spatial position makes them hard to provide adequate information support to ADS. 

\subsection{Vision-Language Model}
Traffic-sign understanding is formulated as an image-to-text task over VLMs in this paper, which avoids a complex training process and plenty of difficult-to-obtain intensive symbol and text data. Recently, a representative model (CLIP)~\cite{radford2021learning} was designed by OpenAI, which trained two encoders via a contrastive learning strategy for the mutual representation of text and image features. BLIP~\cite{li2022blip,li2023blip} proposed to unify vision-language understanding and generation for achieving a wide range of vision-language tasks. MiniGPT4~\cite{zhu2023minigpt} enhanced the model capability to achieve those vision-language tasks by bridging between visual modules and LLMs. LLaVA~\cite{liu2024visual} and LLaVA1.5~\cite{liu2024improved} had further achieved significant progress by equipping themselves with more advanced LLMs. 

Since input image resolution is an important factor for VLMs improving their capabilities to describe images, a lot of researches concentrate on resolution improvement. Qwen-VL~\cite{bai2023qwen} increased the input resolution that supports encoding to 448. Fuyu-8B~\cite{fuyu-8b} adopted different-sized images as pre-training supervision signals, which helped OtterHD~\cite{li2023otterhd} further improve the image resolution. Monkey~\cite{li2024monkey} ensured high resolution by extracting global features from the original image along with fine local details. To achieve a better description for an image with multiple varied objects, GeoChat~\cite{kuckreja2024geochat}, RegionGPT~\cite{guo2024regiongpt}, and ASM~\cite{wang2023all,wang2024all} formulated region-level description as an image-level detailed interpretation. However, such general-domain VLMs perform poorly for first-person view traffic sign scenarios, leading to inaccurate or superfluous and irrelevant information when presented with traffic sign-specific queries. Such a behavior emerges due to the unique challenges introduced by traffic sign interpretation of the vehicle's first-person view.
\begin{figure*}[t]
	\centering
	\includegraphics[width=0.8\linewidth]{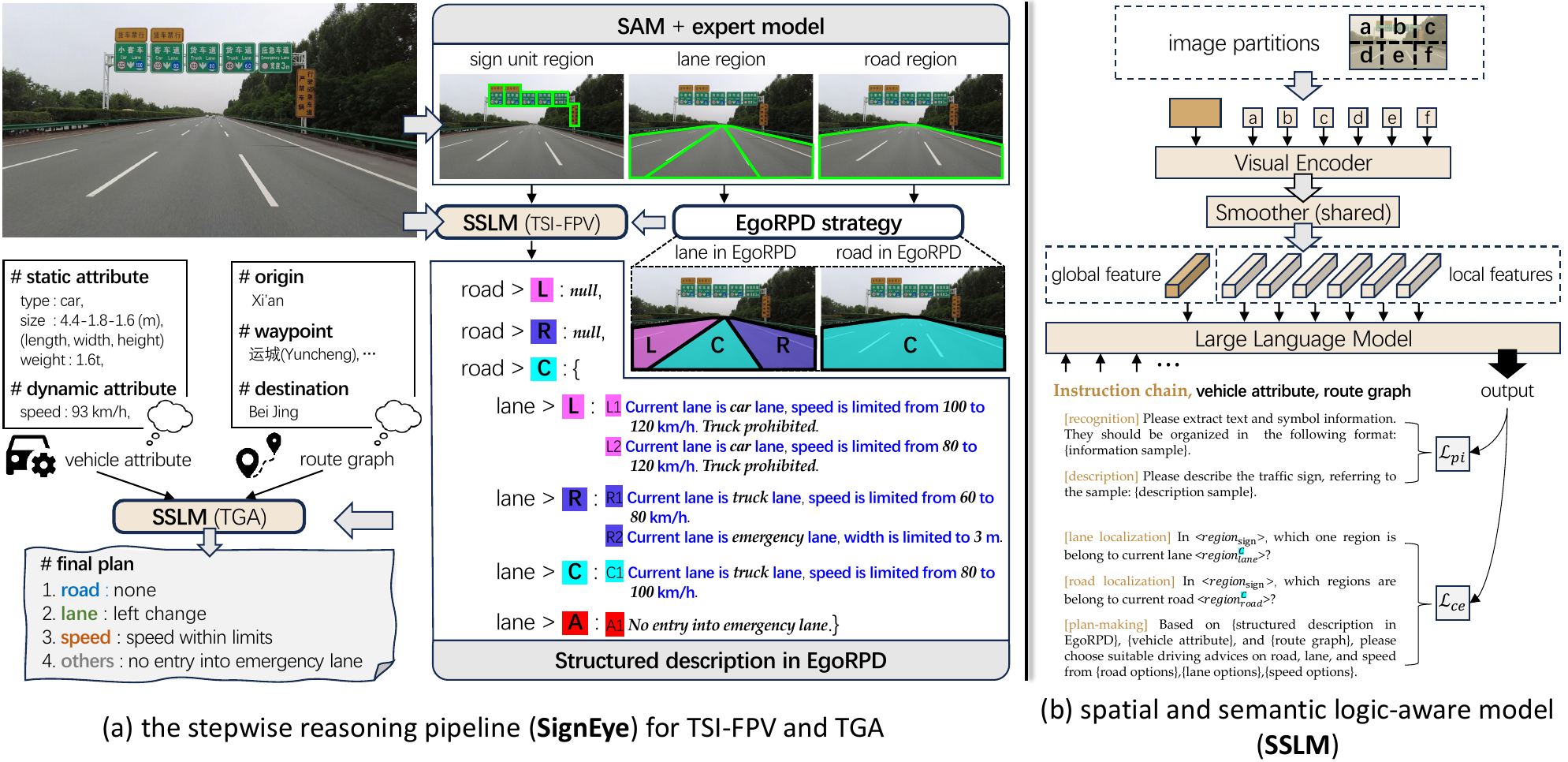}
	\vspace{-.15in}
	\caption{Overall pipeline of SignEye. It takes a road image from the vehicle's first-person view, and object (sign, lane, and road) regions as input to generate structured sign descriptions in EgoRPD and combines them with vehicle attributes and route graphs to achieve the TGA scenario, automatically.}
	\label{fig:V3}
	\vspace{-.15in}
\end{figure*}
\section{Method}
\label{Method}
In this paper, a stepwise reasoning pipeline (SignEye) is constructed to achieve TSI-FPV and TGA. We will describe the details of SignEye in this section.

\subsection{Overall Pipeline of SignEye}
\label{Overall Pipeline}
Existing representative VLMs (such as BLIP2~\cite{li2023blip}, MiniGPT-4~\cite{zhu2023minigpt}, LLaVA~\cite{liu2024visual}, Qwen-VL~\cite{bai2023qwen}, Yi-VL~\cite{young2024yi}, DeepSeek-VL~\cite{lu2024deepseek}, Intern-VL~\cite{chen2024far}, MiniCPM~\cite{yao2024minicpm}, Monkey~\cite{li2024monkey}) focus on pre-training their models on large datasets and aim to achieve a strong ability on general image description. However, the generated description from them is either too long or too short for the introduced TSI-FPV task, which leads to information redundancy or omission and further interferes with the final decision-making about the driving plan. Besides, the understanding of the latent spatial and semantic logic among signs and the position information around the vehicle's first-person view is essential for TGA. However, the large datasets and multiple objective tasks used for pre-training those VLMs do not involve the egocentric vehicle's position information analysis process. As a result, even when these models are well-trained for describing general images, they struggle to accurately transfer traffic sign units to structured descriptions from the first-person view. This limitation affects the models to effectively assist in driving plan decision-making.

To this end, we introduce SignEye. It takes a road image from the vehicle's first-person view, and object (sign, lane, and road) regions as input to generate sign unit descriptions and combines them with vehicle attributes and route graphs to achieve the TGA scenario, automatically. It can assist ADS as a complement to current popular autonomous technologies (such as obstacle perception~\cite{tian2024occ3d,ma2024cotr,huang2024selfocc}) to provide extra traffic-sign-based traffic regulation assessment and direction navigation assistant beyond current popular technologies. Specifically, as shown in Figure~\ref{fig:V3}(a), SignEye adopts several advanced models as components and follows multiple steps to achieve TSI-FPV and TGA:

\textbf{Region proposal generation.} SAM~\cite{kirillov2023segment} and expert models (i.e., SignDet~\cite{yang2024traffic} and UFLD~\cite{qin2020ultra}) are adopted for generating the region proposals of roads, sign units, and lanes, respectively. SAM, a strong segmenter, is finetuned through the road data in RS10K~\cite{guo2023visual} to provide road masks in the pipeline. Considering there are lots of adhesive samples among sign units and lanes while the point or box prompt of each instance is expensive to obtain, the SignDet~\cite{yang2024traffic} and UFLD~\cite{qin2020ultra} are chosen to replace SAM for the responsibility to extract the regions of sign units and lanes in the inference process, where SignDet is trained on our Traffic-CN dataset and UFLD is optimized on both training and test data of Tusimple~\cite{Tusimple2024}.
\begin{table}[t]
	\centering
	\begin{tabular}{c}
		\includegraphics[width=0.8\linewidth]{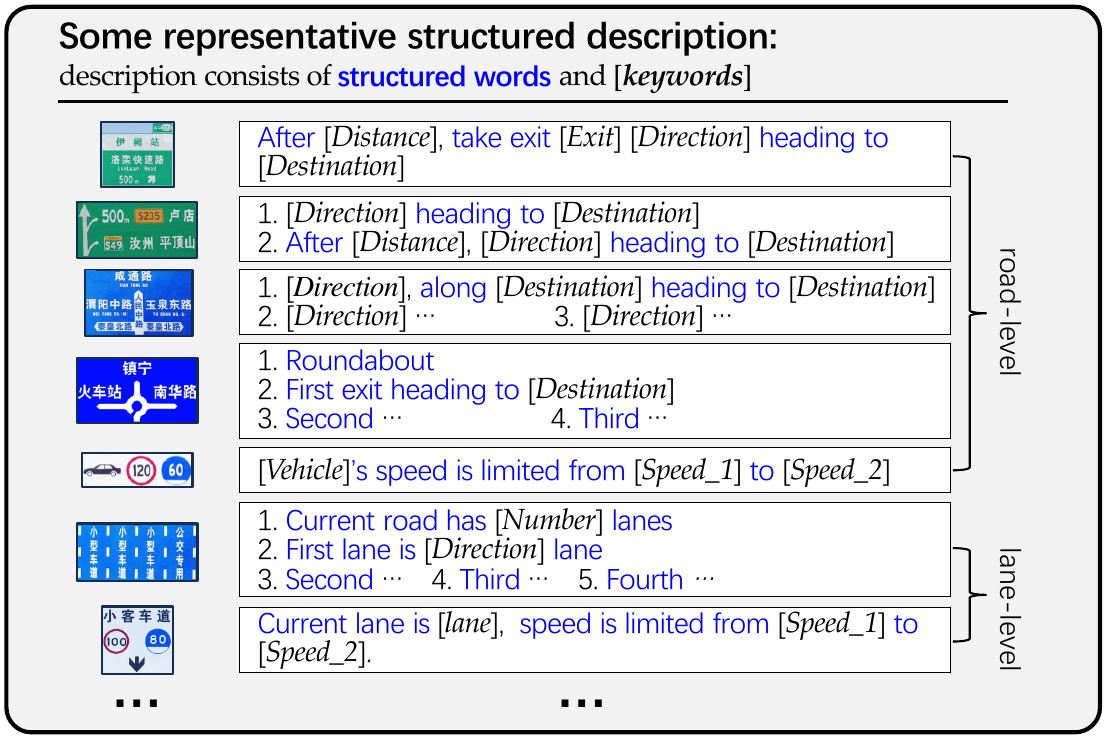}
	\end{tabular}
	\vspace{-.15in}
	\caption{Illustration of structured sign descriptions. Given the space limitations, only a few examples are visualized here.}
	\label{tab:T1}
	\vspace{-.15in}
\end{table}

\textbf{Structured sign description.}
\label{Structured sign description}
Different from formulating traffic sign unit as a directional combination~\cite{guo2021learning} or traffic knowledge graph~\cite{guo2023visual}, we describe it via the structured natural language (as shown in Table~\ref{tab:T1}) according to the~\href{https://openstd.samr.gov.cn/bzgk/gb/newGbInfo?hcno=15B1FC09EE1AE92F1A9EC97BA3C9E451}{Road Traffic Signs and Markings Criteria}. The language description-styled output enjoys intuitional and structured sentence patterns, which makes it easier for the model to take in and further assist in driving plan decision-making for TGA. Importantly, structured descriptions are the key to generating plenty of TGA samples to help our model learn to accomplish traffic regulation assessment and direction navigation (details can be referred to in Section~\ref{Data engine}). Compared with description generation based on text information~\cite{yang2024traffic}, our spatial and semantic logic-aware model (SSLM) combines vision and text features to describe a sign unit, which helps focus on the spatial and semantic logic simultaneously for understanding the traffic symbol and text more effectively. Specifically, SSLM decomposes the describing process into two steps: (1) extracting traffic symbol and text information from the given traffic sign unit and determining the corresponding description structure from Table~\ref{tab:T1} based on information; (2) describing the sign unit following the determined description structure. The two-step process ensures correct spatial semantic logic between different symbols and texts, which either helps prevent the model from generating a redundant description or omitting important information. The details of SSLM can be referred to the following Section~\ref{SSLM Architecture}.
\begin{algorithm}[t]
	\caption{EgoRPD Strategy}  
	\footnotesize
	\begin{algorithmic}[1]
		\Require Points of lane or road lines $\mathbf{S}$ and sign boxes $\mathbf{B}$;
		\Ensure Sign boxes $\mathbf{B}_e$ in EgoRPD;
		\For{$i \gets 1$ to ($N_{S}-1$)}~~//$N_{S}$ \textit{is line number}
		\State	$\theta_{i-1}, \theta_i \gets$ \Call{Arctan(PolyFit}{$\mathbf{S}[i-1]$;$\mathbf{S}[i]$})
		\If{$\theta_{i-1}<=90$ and $\theta_i<=90$} break
		\EndIf  
		\EndFor 
		\State $\mathbf{S}^L_e,\mathbf{S}^C_e,\mathbf{S}^R_e \gets \mathbf{S}[:i],\mathbf{S}[i-1:i+1],\mathbf{S}[i:]$
		\For{$j \gets 0$ to ($N_{B}-1$)}~~//$N_{B}$ \textit{is box number}
		\State	$d^L\gets|\mathbf{B}^{x_{mid}}_{j}- \mathbf{S}^{L,x_{mid}}_{e,\frac{2H}{3}}|$ 
		\State	$d^C\gets|\mathbf{B}^{x_{mid}}_{j,\frac{2H}{3}}- \mathbf{S}^{C,x_{mid}}_{e,\frac{2H}{3}}|$ 
		\State	$d^R\gets|\mathbf{B}^{x_{mid}}_{j,\frac{2H}{3}}- \mathbf{S}^{R,x_{mid}}_{e,\frac{2H}{3}}|$~~//$\mathbf{\cdot }^{x_{mid}}_{\frac{2H}{3}}$ \textit{is the x-axis midpoint of $\mathbf{S}$ two points that y equals $\frac{2}{3}$ image height $H$}
		\If{$d^L=$\Call{Min}{$d^L,d^C,d^R$}} $\mathbf{B}^L_e \gets \mathbf{B}_j$
		\ElsIf{$d^C=$\Call{Min}{$d^L,d^C,d^R$}} $\mathbf{B}^C_e \gets \mathbf{B}_j$
		\ElsIf{$d^R=$\Call{Min}{$d^L,d^C,d^R$}} $\mathbf{B}^R_e \gets \mathbf{B}_j$
		\EndIf~~//$\mathbf{\cdot }^{C}_e$ \textit{is the box assigned to current road/lane }
		\EndFor 
	\end{algorithmic}  
	\label{alg:A1}
\end{algorithm}

\textbf{Structured description in EgoRPD.} Previous sign-related methods did not establish a connection between signs and the egocentric vehicle's position, which results in those methods could not provide effective sign information for ADS. Following the above issues, we introduce the EgoRPD strategy from the vehicle's egocentric view to determining the vehicle's corresponding lane and road positions, which helps connect egocentric vehicles and related signs to support ADS achieve traffic regulation assessment and direction navigation for the current vehicle. Importantly, the EgoRPD strategy only requires considering lanes around the vehicle, which helps avoid the recognition interference brought by perspective distortion of the distant lane and can provide adequate adjacent lane information for single-step lane change process at the same time.


Specifically, the strategy firstly takes lane and road regions as input and assigns them to left (`L'), current (`C'), right (`R'), and all (`A') parts, respectively (referred to Algorithm~\ref{alg:A1}~r1--r6). It then determines whether a sign unit is a lane-level or road-level traffic guidance according to the [\textbf{\textit{keywords}}] of the corresponding structured description from SSLM (as the introduction in Section~\ref{Structured sign description}). Next, SSLM assigns all units to the corresponding road and lane according to their relative position analysis (Algorithm~\ref{alg:A1}~r7--r15). In this stage, SSLM formulates the assignment process as a single-choice question and a multi-choice question for lane and road, respectively, because only one sign unit corresponds to one lane, and there may be multiple units related to one road. It takes the image features, and the coordinate-style sign regions, lane and road regions with `C' mark as input to achieve the the unit assignment (as shown in Figure~\ref{fig:V3}(b) lane and road localization) and output the structured description in EgoRPD. Notably, units belonging to the left or right lane/road can be determined via the position in the X-axis relative to the unit with `C' mark. The trainable ground truth data for this assignment stage is generated via our data engine based on Algorithm~\ref{alg:A1}.
\begin{table}[]
	\renewcommand{\arraystretch}{1}
	\setlength{\tabcolsep}{1.mm}
	\caption{List of options for different-level questions.}
	\vspace{-.15in}
	\centering
	\footnotesize
	\begin{tabular}{ll}
		\toprule
		\rowcolor{grayyy}Level   & Option  \\ \midrule
		road&none; stay; left change; turn left; right change; turn right;\\
		&exit;\\
		lane&none; stay; left change; right change;\\ 
		speed&none; speed within limits; speeding; driving too slowly;\\
		other&none; excessive vehicle height; excessive vehicle width;\\
		&excessive vehicle weight; [description];\\ \bottomrule
	\end{tabular}
	\label{tab:T2}
	\vspace{-.15in}
\end{table}

\textbf{Traffic guidance assistant.} Based on the structured description in EgoRPD, we develop the TGA to re-explore the role of traffic signs in ADS by achieving traffic regulation assessment and direction navigation for the vehicle in the first-person view. In the traffic regulation assessment aspect, vehicle attributes are introduced to simulate the vehicle's driving state, which consists of static (type, size, and weight of vehicle) and dynamic (vehicle speed) attributes. For direction navigation, a route graph is pre-defined and contains plenty of latent waypoints from the origin and destination. With the context information of description, vehicle attributes, and route graph, SSLM formulates the plan generation as a single-choice question. For example, if the current lane limits vehicle speed from 90 to 120 km/h, the SSLM will choose the option of driving too slowly when the speed is 60 km/h. Notably, the `other' plan is mainly responsible for estimating limitations on the vehicle's height, width, and weight. Our model will output the corresponding description directly if there is no information related to the vehicle's height, width, and weight. The pre-defined options for different questions is shown in Table~\ref{tab:T2}.

\subsection{SSLM Architecture}
\label{SSLM Architecture}
Different from previous sign understanding works~\cite{guo2023visual,guo2024signparser,yang2024traffic}, SSLM transfers road images to structured descriptions in EgoRPD, which is essential to achieve TGA for supporting ADS to achieve traffic regulation assessment and direction navigation. SSLM is shown in Figure~\ref{fig:V3}(b), which consists of a visual encoder, smoother, and a LLM-based decoder.

\textbf{Visual encoder.} Specifically, given an input road image $\boldmath{\boldsymbol{I}} \in \mathbb{R}^{H\times W\times 3}$, we divide it into 2$\times$3 slices for support handling the high-resolution (i.e., 1344$\times$1344 resolution) images, where $H$ and $W$ are the height and width of the input image. These slices then are fed into the visual encoder (the pre-trained SigLIP-ViT~\cite{zhai2023sigmoid} is employed as our visual encoder) for embedding them as the corresponding features $\boldmath{\boldsymbol{I}}_v\in \mathbb{R}^{N\times D}$, where $N$ and $D$ are the number of image slices and the feature hidden dimension, respectively.

\textbf{Smoother.} Considering the large hidden dimension $D$ ($D$ is set as 1096 in the pre-trained model) of the feature $\boldmath{\boldsymbol{I}}_v$ from the visual encoder leads to a high token count, which results in extensive GPU memory consumption in the training and inference process, we insert a Smoother after visual encoder to reduce the hidden dimension of $\boldmath{\boldsymbol{I}}_v$ from $D$ to $M$ ($M<<D$). The smoothed visual features $\boldmath{\boldsymbol{I}}_s \in \mathbb{R}^{N\times M}$ allows us to train and evaluate the model on RTX 4090 of 24 GB memory and achieves competitive results.

\textbf{LLM-based decoder.} To decode the combination features of vision and text into the natural language description, a strong open source LLM (QWen2~\cite{qwen2}) is adopted as our decoder. It takes a sequence of smoothed visual features $\boldmath{\boldsymbol{I}}_s$ and pre-designed instruction tokens $\boldmath{\boldsymbol{T}}_{ins}$ as input, generating task-specific answers.

\subsection{SSLM Training}
We employ a strategy that involves initializing the visual encoder and LLM-based decoder with pre-trained weights proposed in SigLIP~\cite{zhai2023sigmoid} and QWen2~\cite{qwen2} and fine-tuning specific segments with the Low-Rank Adaptation (LoRA~\cite{hu2021lora}) method for SSLM. In the fine-tuning process, considering even simple descriptions (e.g., examples in Table~\ref{tab:T1}), VLMs often struggle to generate a structured description due to the hallucination problem. Meanwhile, VLMs hard to propose suitable advice for the TGA that is dependent on spatial relative position deeply. To this end, we follow the stepwise pipeline to achieve multi-task instruction tuning (as shown in Figure~\ref{fig:V3}(b)).

For the SSLM (TSI-FPV) (the extraction and description tasks in Figure~\ref{fig:V3}(a)) optimization, we formulate the process as maximizing a permutation-invariant likelihood function $\mathcal{L}_{pi}$ to estimate the difference between the input answer and the corresponding ground truth, which avoids the interference brought by the arrangement sequence of [\textbf{\textit{keywords}}]. Specifically, as shown in Table~\ref{tab:T1}, there always more than one keywords within the same [\textbf{\textit{keywords}}] position (e.g., \textcolor{blue}{\textbf{heading to}} [\textbf{\textit{Destination1, Destination2, ...}}]). The sign unit description is correct no matter how the relative position between \textbf{\textit{Destination1}} and \textbf{\textit{Destination2}} distribute. The permutation-invariant likelihood function is formulated as:
\begin{gather}
\mathcal{L}_{pi}=\mathrm{min}(\mathrm{log}P(\boldmath{\boldsymbol{T}}^k_{ans}|\boldmath{\boldsymbol{I}}_s,\boldmath{\boldsymbol{T}}_{ins};\Theta)),\\
k=1,2,...,M!\times \prod_{j}^{M} V_j!,
\label{eqn:E1}
\vspace{-.25in}
\end{gather}
where $\boldmath{\boldsymbol{T}}_{ins}$ and $\Theta$ are the input instruction tokens and the trainable parameters. $P$ is the conditional probability. $M$ represents the number of sub-sentences, and $V$ denotes the number of [\textbf{\textit{keywords}}] in each sub-sentence. $\boldmath{\boldsymbol{T}}^k_{ans}$ denotes the $k$ th styled answer tokens, which is generated through the re-arrangement of sub-sentences and [\textbf{\textit{keywords}}] sequence of the given answer tokens $\boldmath{\boldsymbol{T}}_{ans}$. 

For the SSLM (TGA) (the localization and plan-making tasks in Figure~\ref{fig:V3}(a)) optimization, we estimate the prediction via cross entropy function directly.

\begin{figure}[t]
	\centering
	\includegraphics[width=0.8\linewidth]{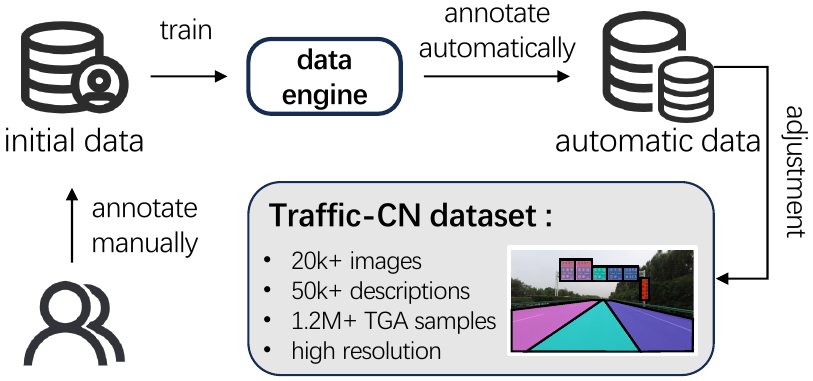}
	\vspace{-.1in}
	\caption{Workflow of the data engine for building the Traffic-CN.}
	\label{fig:V4}
	\vspace{-.15in}
\end{figure}
\section{Traffic-CN Dataset}
\label{Traffic-CN Dataset}
\textbf{Images.} Traffic-CN collects 20k+ images with high resolution (1920$\times$1080 pixels). They are captured from the vehicle's first-person view through the DJI OSMO Pocket2 camera. The dataset involves some popular Chinese provinces (including Shaanxi, Henan, Shanxi, Hebei, Tianjin, and Beijing), typical road scenes (such as highways, urban roads, urban streets, and rural roads), and various challenged optical environments (e.g., overexposure, rain and fog blur, shadows, occlusion, and motion blur).

\textbf{Data engine.}
\label{Data engine}
To fulfill the research of the TSI-FPV and TGA, we build a data engine (Figure~\ref{fig:V4}) to enable the label collection of the Traffic-CN dataset. For \textbf{TSI-FPV data}, the data engine follows three stages to generate structured descriptions: (1) a totally manual labeling stage for a small amount of initial data; (2) a model-assisted automatic annotation stage, where SignEye is trained on initial data to predict automatic annotations; (3) a manual adjustment stage, where partial automatic data is corrected manually and used for training model again with initial data together. In the end, repeat stage (3) for labeling all image sign units. Our data engine produced 50k+ structured descriptions for sign units, 85\% of which were generated fully automatically. \textbf{TGA data} consists of localization and plan-making data (as shown in Figure~\ref{fig:V3}(b)). The former is generated according to the spatial distribution among sign, lane, and road regions through Algorithm~\ref{alg:A1}. The plan-making data is generated by combining structured descriptions in EgoRPD from the localization task with vehicle attributes and the route graph, where the description is decomposed into smaller traffic instructions following its structure, and vehicle attributes and waypoints are matched with the instruction for generating suitable advice from options. Based on the images, the introduced data engine generate 1.2M+ TGA data automatically.


\section{Experiments}
\begin{figure*}[t]
	\centering
	\includegraphics[width=0.7\linewidth]{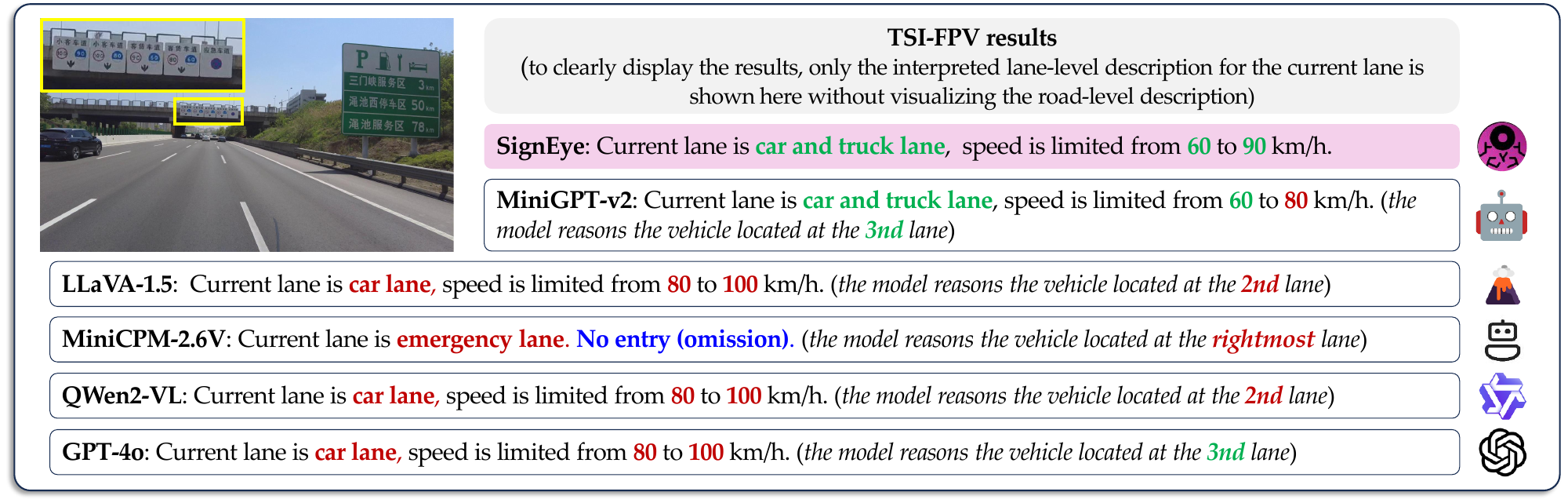}
	\vspace{-.15in}
	\caption{Visualization of the egocentric vehicle corresponding description in the TSI-FPV task.}
	\label{fig:V5}
	\vspace{-.15in}
\end{figure*}
\subsection{Implementation Details}
We initialize our model using the well-trained SigLIP-ViT~\cite{zhai2023sigmoid} and QWen2~\cite{qwen2}. During the training process, we refine the parameters by employing LoRA~\cite{hu2021lora} for low-rank adaptation, with the rank $r$ set to 64. We utilize the AdamW~\cite{loshchilov2017decoupled} optimizer along with a cosine learning rate scheduler to train SSLM over 5 epochs using a batch size of 16 across eight NVIDIA 4090 GPUs.

\subsection{Main Results}
\textbf{Traffic Sign Interpretation (TSI) Task.} This task is introduced to generate accurate structured descriptions to help the model understand the traffic instruction information from sign units more easily. The quality of descriptions influences TGA scenario application directly. We present our model's performance on the Traffic-CN dataset. The results in Table~\ref{tab:T3} indicate that the instruction chain of extraction and description (Figure~\ref{fig:V3}(b)) of SignEye demonstrates enhanced performance on this task compared with general VLMs describing sign units directly.
\begin{table}[t]
	\renewcommand{\arraystretch}{.8}
	\setlength{\tabcolsep}{.5mm}
	\caption{Results on TSI task. `B-1', `B-2', `B-3', and `B-4' are `BLEU-1', `BLEU-2', `BLEU-3', and `BLEU-4'. `R-L' denotes `ROUGE-L' metric.}
	\vspace{-.15in}
	\centering
	\footnotesize
	\begin{tabular}{lccccccc}
		\toprule
		\rowcolor{grayyy}Model   & B-1 & B-2 & B-3 & B-4 & METEOR & R-L & CIDEr \\ \midrule
		MiniGPT-v2&    56.1    &   41.6    &   32.9    &   27.5    &  33.9     &    55.7     &    1.97    \\
		LLaVA-1.5&    73.1    &   64.6    &    58.7   &    53.9   &    45.9    &    71.0     &   4.20    \\ 
		MiniCPM-2.6&   69.4    &   60.6    &  54.0     &    48.6   &    45.6    &     68.6    &   4.05    \\ 
		QWen2-VL&    66.2    &   57.8    &   51.2    &   45.4    &    47.1    &    70.1     &    4.24   \\ \midrule
		\textbf{SignEye} &   \textbf{74.8}     &    \textbf{67.1}   &   \textbf{61.1}    &   \textbf{56.2}    &    \textbf{49.3}    &    \textbf{72.7}     & \textbf{4.70}\\ \bottomrule
	\end{tabular}
	\label{tab:T3}
	\vspace{-.15in}
\end{table}

\textbf{TSI-FPV based TGA scenario application.} To re-explore the role of traffic signs in ADS, we develop a TGA scenario (as visualized at the bottom of Figure~\ref{fig:V1}). Considering the deep dependency of TGA on egocentric vehicle position information, we introduce the EgoRPD strategy to build a connection between vehicle and sign units for further proposing the TSI-FPV task based on previous TSI to support TGA. In Figure~\ref{fig:V5}, we first visualize the model's ability to match the egocentric vehicle and the corresponding sign units. It can be observed that general VLMs struggle to accurately locate the vehicle's position and connect the egocentric view of the vehicle to the corresponding sign units, even though humans can easily interpret these spatial relationships. While GPT-4o and MiniGPT-v2 correctly identify the lane position, it still finds it challenging to associate the position with the corresponding correct lane-level sign unit. Unlike these general VLMs, which determine all lanes sequentially from left to right before identifying the vehicle’s lane, SignEye determines the current lane by analyzing the geometry of lane lines and connects the egocentric view with the corresponding sign unit based on their spatial relationship (see Algorithm~\ref{alg:A1}) rather than by analyzing the absolute spatial location of each instance globally, which brings significant improvements for our model in making lane-level plan (referred to Section~\ref{Effectiveness of the TSI-FPV on TGA}).

To verify the importance of TSI-FPV for TGA, we then show the performance results in Table~\ref{tab:T4}, where SignEye achieves remarkable proficiency in plan accuracy, surpassing the nearest competing general VLMs without descriptions in EgoRPD 7.3\% at least. Importantly, our model achieves 9.7\% superiority when we drop the \#1 option that does not require the position information but only sign units. These results highlight the effectiveness of the connection between the egocentric vehicle and the corresponding sign descriptions provided by TSI-FPV and the stepwise reasoning pipeline. The effectiveness of SignEye for analyzing the complex spatial relationship among the egocentric vehicle, signs, lanes, and roads is also demonstrated.
\begin{table}[t]
	\renewcommand{\arraystretch}{.8}
	\setlength{\tabcolsep}{.9mm}
	\caption{Accuracy results on TGA scenario application. $O^{drop \#1}_{all}$ is the all overall accuracy without \#1 option.}
	\vspace{-.15in}
	\centering
	\footnotesize
	\begin{tabular}{lcccccc}
		\toprule
		\rowcolor{grayyy}Model   & Road & Lane & Speed & Other & $\mathrm{O_{all}}$  &$\mathrm{O^{drop\#1}_{all}}$\\ \midrule
		MiniGPT-v2&   42.6     &    37.4   &    62.2   &   49.5    &   47.9  & 34.0 \\
		LLaVA-1.5&    65.8    &    47.2   &   67.7    &    72.2    &   63.2 & 54.8\\ 
		MiniCPM-2.6&  79.7 &   83.0    &   80.5   &     82.1    &  79.7&  75.9\\ 
		QWen2-VL&   78.8     &   63.8    &   70.5    &    72.6   &   71.5  &  65.2\\ \midrule
		\textbf{SignEye} &    \textbf{88.8}   &  \textbf{86.3}    &    \textbf{85.6}   &  \textbf{87.4}  & \textbf{87.0} &\underline{\textbf{85.6}(\textcolor{greenn}{\textbf{9.7}$\bm\uparrow $})}\\ \bottomrule
	\end{tabular}
	\label{tab:T4}
	\vspace{-.15in}
\end{table}

\begin{table*}[t]
	\renewcommand{\arraystretch}{.5}
	\setlength{\tabcolsep}{.5mm}
	\caption{Ablation study on EgoRPD strategy-based TSI-FPV for TGA. \#n means the $n$th option, and $\mathrm{O_r}$, $\mathrm{O_l}$, $\mathrm{O_s}$ and $\mathrm{O_o}$ are the overall accuracy of all options of road, lane, speed, and other respectively. $\mathrm{O_{all}}$ is the overall accuracy of all options of road, lane, speed, and other. $O^{drop \#1}_{all}$ is the all overall accuracy without \#1 option. `w/o EgoRPD' denotes TSI, and `with EgoRPD' means TSI-FPV.}
	\vspace{-.15in}
	\centering
	\footnotesize
	\begin{tabular}{cccccccccccccc}
		\toprule
		\multirow{2}{*}{EgoRPD} & \multicolumn{7}{c}{\cellcolor{grayyy}Road} & \multicolumn{1}{c|}{\cellcolor{grayyy}$\mathrm{O_r}$} & \multicolumn{4}{c}{\cellcolor{grayyy}Lane}                                & \cellcolor{grayyy}$\mathrm{O_l}$ \\ 
		& \multicolumn{1}{c}{\#1}  & \multicolumn{1}{c}{\#2}  & \multicolumn{1}{c}{\#3}  & \multicolumn{1}{c}{\#4}  & \multicolumn{1}{c}{\#5}  & \multicolumn{1}{c}{\#6}  & \multicolumn{1}{c}{\#7}  & \multicolumn{1}{c|}{/}    & \multicolumn{1}{c}{\#1}  & \multicolumn{1}{c}{\#2}  & \multicolumn{1}{c}{\#3}  & \multicolumn{1}{c}{\#4}  & /    \\ \midrule
		w/o                     & \multicolumn{1}{c}{90.4}   & \multicolumn{1}{c}{85.7} & \multicolumn{1}{c}{62.5} & \multicolumn{1}{c}{70.0}   & \multicolumn{1}{c}{82.1} & \multicolumn{1}{c}{75.0}   & \multicolumn{1}{c}{71.9} & \multicolumn{1}{c|}{86.5} & \multicolumn{1}{c}{90.1}   & \multicolumn{1}{c}{87.8} & \multicolumn{1}{c}{53.3} & \multicolumn{1}{c}{46.4} & 82.6 \\ 
		with                    & \multicolumn{1}{c}{90.2}   & \multicolumn{1}{c}{88.7} & \multicolumn{1}{c}{76.5} & \multicolumn{1}{c}{66.7} & \multicolumn{1}{c}{81.8} & \multicolumn{1}{c}{70.2}   & \multicolumn{1}{c}{88.9} & \multicolumn{1}{c|}{\underline{\textbf{88.8}(\textcolor{greenn}{\textbf{2.3}$\bm\uparrow $})}} & \multicolumn{1}{c}{89.8}   & \multicolumn{1}{c}{88.4} & \multicolumn{1}{c}{{83.3}(\textcolor{red}{\textbf{30.0}}\textcolor{red}{$\bm\uparrow $})} & \multicolumn{1}{c}{{67.9}(\textcolor{red}{\textbf{21.5}}\textcolor{red}{$\bm\uparrow $})} &\underline{\textbf{86.3}(\textcolor{greenn}{\textbf{3.7}$\bm\uparrow $})} \\ \midrule
		\multirow{2}{*}{EgoRPD} & \multicolumn{4}{c}{\cellcolor{grayyy}Speed}                                                                                    & \multicolumn{1}{c|}{\cellcolor{grayyy}$\mathrm{O_s}$} & \multicolumn{4}{c}{\cellcolor{grayyy}Other}                                                                                    & \multicolumn{1}{c|}{\cellcolor{grayyy}$\mathrm{O_o}$} & \multicolumn{1}{c}{\cellcolor{grayyy}$\mathrm{O_{all}}$} & \multicolumn{2}{c}{\cellcolor{grayyy}$\mathrm{O^{drop \#1}_{all}}$}      \\ 
		& \multicolumn{1}{c}{\#1}  & \multicolumn{1}{c}{\#2}  & \multicolumn{1}{c}{\#3}  & \multicolumn{1}{c}{\#4}  & \multicolumn{1}{c|}{/}    & \multicolumn{1}{c}{\#1}  & \multicolumn{1}{c}{\#2}  & \multicolumn{1}{c}{\#3}  & \multicolumn{1}{c}{\#4}  & \multicolumn{1}{c|}{/}    & \multicolumn{1}{c}{/}    & \multicolumn{2}{c}{/}           \\ \midrule
		w/o                     & \multicolumn{1}{c}{87.4} & \multicolumn{1}{c}{67.2} & \multicolumn{1}{c}{78.2} & \multicolumn{1}{c}{65.5} & \multicolumn{1}{c|}{81.5} & \multicolumn{1}{c}{88.5} & \multicolumn{1}{c}{81.3} & \multicolumn{1}{c}{80.1} & \multicolumn{1}{c}{87.3} & \multicolumn{1}{c|}{85.3} & \multicolumn{1}{c}{84.0} & \multicolumn{2}{c}{80.3}        \\ 
		with                    & \multicolumn{1}{c}{88.3} & \multicolumn{1}{c}{82.5} & \multicolumn{1}{c}{84.4} & \multicolumn{1}{c}{77.2} & \multicolumn{1}{c|}{\underline{\textbf{85.6}(\textcolor{greenn}{\textbf{4.1}$\bm\uparrow $})}} & \multicolumn{1}{c}{89.2} & \multicolumn{1}{c}{80.0}   & \multicolumn{1}{c}{84.4} & \multicolumn{1}{c}{90.8} & \multicolumn{1}{c|}{\underline{\textbf{87.4}(\textcolor{greenn}{\textbf{2.1}$\bm\uparrow $})}} & \multicolumn{1}{c}{\underline{\textbf{87.0}(\textcolor{greenn}{\textbf{3.0}$\bm\uparrow $})}} & \multicolumn{2}{c}{\underline{\textbf{85.6}(\textcolor{greenn}{\textbf{5.3}$\bm\uparrow $})}}        \\ \bottomrule
	\end{tabular}
	\label{tab:T5}
	\vspace{-.15in}
\end{table*}
\subsection{Ablation Studies}
We conduct thorough experiments to validate the effectiveness of the designed stepwise reasoning pipeline and permutation-invariant loss.

\textbf{Effectiveness of the TSI-FPV on TGA.}
\label{Effectiveness of the TSI-FPV on TGA}
As shown in Figure~\ref{fig:V3}, the vehicle's relative position information is important for TGA in making a driving plan, especially for lane and speed plans. We visualize the plan accuracy of road, lane, speed, and other in Table~\ref{tab:T5} to show the effectiveness of the EgoRPD strategy-based TSI-FPV task on TGA. 

Specifically, for all kinds of plans, SignEye achieves superior accuracy in \#1 option. It is mainly because \#1 corresponds to the `none' option (referred to~\ref{tab:T2}), which means the model can choose the correct option by recognizing whether signs occurred in images and no need for position information. Instead, for those relative position-sensitive lane change options (e.g., lane \#3 and \#4), the model has to determine the lane-level sign description corresponding to the egocentric vehicle's lane position first. Then, it finds out the adjacent left and right lane-level sign descriptions. In the end, SignEye makes a lane plan according to vehicle attributes and descriptions. It can be observed that our model achieves 30\% and 21\% improvements on the lane change options compared with the model that is without the description in EgoRPD and has to analyze the complex spatial distribution based on vision features only. A similar conclusion can be given for making a speed plan because there are plenty of situations in which the speed description occurs on lane-level sign units. We also show the overall accuracy of all options of road, lane, speed, and other, the EgoRPD strategy can bring 3.0\% and 5.3\% improvements when with and without considering the `none' option. The results demonstrate the essential of EgoRPD strategy to TGA, especially for position-sensitive plans.

\textbf{Effectiveness of Permutation-Invariant Loss on TSI.} As described in Table~\ref{tab:T1} and Section~\ref{eqn:E1}, the prediction description always corresponds to multiple order-adjusted ground truths for samples with multiple sub-sentences and destinations. The designed $\mathcal{L}_{pi}$ optimizes the model more efficiently by minimizing the gap between the prediction and the most matched ground truth among the multiple options. It brings performance improvements to the TSI task as the number of [\textit{\textbf{keywords}}] in the description increases, as demonstrated in Table~\ref{tab:T6}.

\section{Discussion and Conclusion}
\subsection{Discussion}
\textbf{TSI-FPV \textit{vs.} Traffic Sign Understanding.} Traffic signs always corresponding to different located roads or lanes, making it essential to identify signs relevant to the egocentric vehicle based on road and lane positions. Unlike previous methods that ignore position information, TSI-FPV enables ADS to apply relevant signs for egocentric driving plans and avoid interference from unrelated signs.
\begin{table}[t]
	\renewcommand{\arraystretch}{.8}
	\setlength{\tabcolsep}{1.mm}
	\caption{Performance gain brought by $\mathcal{L}_{pi}$ to the TSI task. $\mathrm{num_{k}}$ denotes the number of [\textit{\textbf{keywords}}] in a ground truth description.}
	\vspace{-.15in}
	\centering
	\footnotesize
	\begin{tabular}{lccccccc}
		\toprule
		\rowcolor{grayyy} $\mathrm{num_{k}}$  & B-1 & B-2 & B-3 & B-4 & METEOR & R-L & CIDEr \\ \midrule
		1&   {{0.0~~~~~}\textcolor{greenn}{$\bm$}}     &    {{0.0~~~~~}\textcolor{greenn}{$\bm$}}   &    {{0.0~~~~~}\textcolor{greenn}{$\bm$}}   &    {{0.0~~~~~}\textcolor{greenn}{$\bm$}}   &   {{0.0~~~~~}\textcolor{greenn}{$\bm$}}    &    {{0.0~~~~~}\textcolor{greenn}{$\bm$}}     &   {{0.0~~~~~}\textcolor{greenn}{$\bm$}}    \\
		4&   {\textbf{0.8}\textcolor{greenn}{$\bm\uparrow$}}     &    {\textbf{1.4}\textcolor{greenn}{$\bm\uparrow$}}   &    {\textbf{1.5}\textcolor{greenn}{$\bm\uparrow$}}   &    {\textbf{1.3}\textcolor{greenn}{$\bm\uparrow$}}   &   {\textbf{0.1}\textcolor{greenn}{$\bm\uparrow$}}    &    {\textbf{0.8}\textcolor{greenn}{$\bm\uparrow$}}     &   {\textbf{0.08}\textcolor{greenn}{$\bm\uparrow$}}    \\
		7&   {\textbf{2.4}\textcolor{greenn}{$\bm\uparrow$}}    &   {\textbf{4.3}\textcolor{greenn}{$\bm\uparrow$}}  &   {\textbf{5.6}\textcolor{greenn}{$\bm\uparrow$}}    &   {\textbf{6.4}\textcolor{greenn}{$\bm\uparrow$}}    &   {\textbf{0.5}\textcolor{greenn}{$\bm\uparrow$}}    &   {\textbf{3.8}\textcolor{greenn}{$\bm\uparrow$}}     & {\textbf{0.51}\textcolor{greenn}{$\bm\uparrow$}}  \\  \bottomrule
	\end{tabular}
	\label{tab:T6}
	\includegraphics[width=0.9\linewidth]{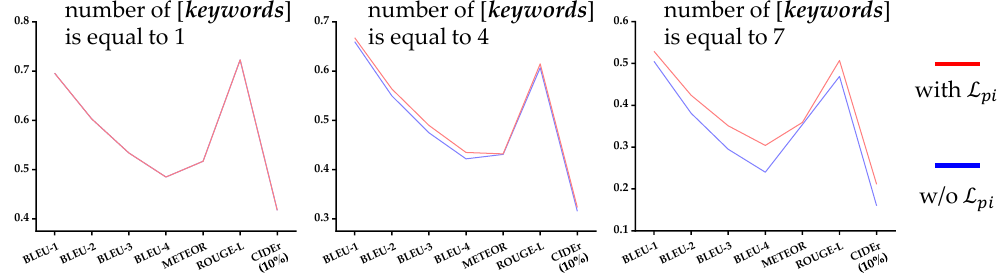}
	\vspace{-.15in}
\end{table}

\textbf{Structured Description \textit{vs.} General Description.} Unlike general description tasks (e.g., image captioning, VQA), SignEye generates structured descriptions (see Table~\ref{tab:T1}) without redundancy or omissions, which offers two advantages: (1) it enables VLMs to convey key information directly to ADS or drivers more effectively, and (2) it allows our data engine to perform regularized analysis of key information in sign units for generating large-scale TGA data.

\textbf{TGA \textit{vs.} Electric Map Navigation.} Both of them are responsible for direction navigation and traffic regulation assessment. However, \textit{\textbf{they are not substitutive but complementary}}. TGA enjoys the superiority of dynamic and offline relative to electric map navigation, which makes them complementary: (1) TGA can provide newly added or changed sign information for ADS to handle the situations of road construction or rule's temporary adjustment; (2) the TSI-FPV of TGA can maintenance of the sign information for the electric map as an automatic tool; (3) TGA complements the electric map under weak signal or offline.

\subsection{Conclusion}
In this paper, we introduce TSI-FPV to interpret sign units as structured descriptions in EgoRPD. Building on it, we develop TGA to re-explore the role of traffic signs in assisting ADS. Additionally, the Traffic-CN dataset, created through our data engine, supports research and evaluation of TSI-FPV and TGA, encouraging broader participation in sign-related ADS research. Experiments show that TSI-FPV and TGA are feasible, and the SignEye achieves superior performance over general VLMs. In future work, we will focus on integrating TGA with mainstream ADS technologies (e.g., occupancy networks) to enhance sign-related support for driving plan-decisions.
\newpage

{
    \small
    \bibliographystyle{ieeenat_fullname}
    \bibliography{main}
}
\end{document}